\documentclass[runningheads]{llncs}
\usepackage[T1]{fontenc}
\usepackage{booktabs, multirow}
\usepackage{graphicx,verbatim}
\usepackage{amsmath}
\usepackage{textgreek}
\usepackage{hyperref}
\hypersetup{colorlinks=true, linkcolor=blue, citecolor=blue, urlcolor=blue}

\begin{document}
\title{ Does FLAIR super-resolution erase or hallucinate small white-matter lesions?}
\titlerunning{Effects of Super-Resolution on WMH Detection}
%

\author{Zahra Khodakarami, Yue Li, Pulkit Khandelwal, John Detre, Sandhitsu Das, Christopher Brown, David Wolk, Paul Yushkevich}  
\authorrunning{Z. Khodakarami et al.}
\institute{University of Pennsylvania, Philadelphia, PA, USA \\
    \email{zkm@engineering.upenn.edu}}
  
\maketitle              
\begin{abstract}
White matter hyperintensities (WMH), bright regions on
Fluid-attenuated Inversion Recovery (FLAIR) scans are associated with cerebrovascular pathology and neurodegeneration. FLAIR is usually acquired with thick slices in clinical settings, giving it poor through-plane resolution. Super-resolution (SR) is a widely used method for recovering an isotropic volume from an anisotropic scan. Yet whether applying it prior to WMH segmentation preserves lesion content remains unknown: a model may erase small real lesions or hallucinate absent ones.
We used 1-mm isotropic high-resolution (HR) FLAIR scans from 29 individuals in the ADNI cohort, each manually segmented for WMH by an expert. Then, we degraded each to simulated 3 and 5 mm through-plane acquisitions. Multi-contrast implicit neural representation (INR), a single-contrast self-supervised model (ECLARE), and cubic interpolation were used to upsample them onto the HR grid. WMH segmentation from a simulated thick slice and the original HR FLAIR set the floor and ceiling, respectively, for the per-lesion analysis. Of four WMH segmentation methods (WMH-SynthSeg, segcsvd, MARS-WMH, TrUE-Net), we ran the analysis under the most sensitive one to small lesions on HR (MARS-WMH) with the evaluation metrics of detection sensitivity, erasure rate (HR-detected lesions lost after reconstruction), and hallucination rate (predicted components absent from both the manual and HR segmentation). The dominant effect of SR was erasure of small real lesions, not hallucination, and it increased with slice thickness, though every reconstruction still improved lesion detection over the raw thick slice. ECLARE recovered small lesion signal best at both thicknesses, while the INR was no better than cubic interpolation.

\keywords{WMH Segmentation \and Super-Resolution \and Anisotropic Scans.}

\end{abstract}
\section{Introduction}

White matter hyperintensities (WMH) are the primary MRI marker of cerebral small vessel disease (CSVD)~\cite{strive2013} and are a well-established neuroimaging biomarker that predicts future stroke, cognitive decline, and dementia~\cite{debette2010}. Measuring the total volume of WMH has long been important in clinical practice and research. Hyperintense lesion load is a standard trial endpoint in multiple sclerosis and other demyelinating diseases~\cite{filippi1998,sormani2013}, and it is equally important in CSVD, where changes in WMH volume are used as a neuroimaging outcome in clinical trials~\cite{csvdtrialB1}. Automated WMH segmentation methods label every lesion voxel to calculate total lesion volume. This quantification is objective, reproducible, and scalable~\cite{cognitionA1,cognitionA2}. High sensitivity to subtle changes is crucial, particularly for the small early lesions where disease progression first becomes visible.

Sensitivity of WMH detection, however, depends critically on the quality of the underlying acquisition. Fluid-attenuated inversion recovery (FLAIR) is the reference sequence for WMH, suppressing CSF signals so that lesions are distinct bright foci within white matter~\cite{strive2013}. Resolving small lesions requires near-isotropic, high-resolution sampling, but usually scan-time budgets in clinical practice enforce thick-slice, 2D/anisotropic FLAIR with poor through-plane resolution. This leads to partial-volume averaging across thick slices, which can blur or even obscure small lesions, and information lost at acquisition cannot be recovered downstream. Guo et al. quantified the impact of slice thickness on reliability, reporting 2D versus 3D FLAIR WMH volume disagreement of 15.6--29.3\% CV compared with around 3\% for repeated 3D acquisitions, thought to be attributable to slice thickness~\cite{guo2019}. 

One strategy to mitigate this loss of through-plane information is super-resolution (SR), which upsamples anisotropic FLAIR into an isotropic volume prior to segmentation. The simplest form is polynomial interpolation, which redistributes observed intensities but adds no new anatomical detail. Learning-based methods instead draw on trained priors or complementary modalities, such as a T1-weighted (T1w) scan. Single-contrast self-supervised SR, such as ECLARE~\cite{remedios2024} and its predecessor SMORE~\cite{zhao_smore}, reconstructs the through-plane dimension from the FLAIR scan alone. Conversely, multi-contrast approaches, such as implicit neural representation (INR)-based methods~\cite{mcginnis2023}, exploit the co-acquired isotropic T1w as an anatomical prior to guide upsampling.

Prior work has shown that SR can improve downstream lesion segmentation relative to naive interpolation, as measured by spatial overlap metrics such as Dice~\cite{li2022}. However, overlap metrics are mostly dominated by large lesions that account for most of the WMH volume and are less sensitive to errors in smaller lesions. Two distinct failure modes may occur: SR may \emph{erase} a small real lesion, smoothing it back into the white matter, or, because these models are optimized for perceptual fidelity, it may \emph{hallucinate} a plausible-looking lesion where none exists~\cite{bhadra2021,muckley2021}. The first depresses sensitivity; the second inflates the count with false positives. Neither is captured in an aggregate overlap score, and to date, neither has been characterized at the level of the individual lesion.

We address this gap directly by evaluating SR not only by perceptual quality or bulk overlap, but also by its impact on individual lesions: the true lesions it preserves versus the spurious ones it introduces. Using a controlled degradation of native-isotropic FLAIR as a true high-resolution reference, our contributions are threefold: (1) we demonstrate that automated WMH segmenters differ widely in their sensitivity to small lesions on HR FLAIR, we therefore adopt the most sensitive segmenter, MARS-WMH, as the primary reader of reconstruction; (2) we benchmark ECLARE against a T1-guided INR SR and cubic interpolation, quantifying both per lesion and by lesion size how much lesion signal each recovers across 3 and 5 mm through-plane thicknesses; and (3) we decompose the per-lesion cost of reconstruction into the erasure of HR-detected lesions (which is concentrated among the smallest lesions) and the hallucination rate (reconstruction-induced false-positive components matching neither the manual mask nor the segmenter's HR segmentation).

\section{Materials and Methods}
\subsubsection{Dataset.}
We study WMH in an older adult cohort of 29 individuals from ADNI, whose WMH masks were delineated by an expert on the 1 mm-isotropic HR FLAIR, providing the lesion-level reference. Connected-component analysis of the manual masks yields 2{,}597 distinct reference lesions (2{,}016 <3 mm, 468 3--6 mm, 74 6--10 mm, and 39 >10 mm by equivalent diameter), serving as the fundamental unit for all lesion-level analyses. From each HR scan, we synthesize a thick-slice low-resolution (LR) FLAIR by blurring the superior-inferior (through-plane) axis with a Gaussian slice-profile kernel (FWHM equal to the slice thickness) and downsampling to the target spacing. This leaves the in-plane resolution untouched, simulating the thick axial 2D FLAIR used in clinical practice. We assessed two representative thicknesses (3 and 5 mm). The HR FLAIR is used only for evaluation and never for reconstruction~\cite{mcginnis2023,remedios2024}. After skull-stripping, N4 bias-field correction, and rigid registration of the T1w into 1 mm HR-FLAIR space, both imaging contrasts and the manual mask lie on one grid.

\subsection{Super-resolution Methods}
We evaluate three reconstructions of the simulated thick-slice FLAIR: multi-contrast INR, single-contrast ECLARE, and cubic B-spline interpolation, all matched to the HR grid prior to segmentation. The native HR FLAIR is the ceiling; the floor is the thick-slice LR FLAIR segmented directly, its labels nearest-neighbor–resampled to the HR grid for comparison.
\subsubsection{Multi-contrast INR.} Following McGinnis et al.~\cite{mcginnis2023}, this network uses the co-registered, native-resolution T1w to guide FLAIR reconstruction. The T1w retained its native 1 mm resolution and was co-registered with the FLAIR (resliced onto the 1 mm HR grid), providing the strongest anatomical prior, so any WMH erasure or hallucination cannot be attributed to degraded guidance.
\subsubsection{Single-contrast ECLARE.} Run in its default configuration~\cite{remedios2024}, this self-supervised method estimates the slice profile via ESPRESO algorithm~\cite{espreso} to recover through-plane resolution using only the LR FLAIR.
\subsubsection{Cubic interpolation.} A non-learning baseline that preserves the native FLAIR intensities. The learned models (INR, ECLARE) instead output normalized intensities, which we rescale to the HR FLAIR range (1st–99th within-brain percentiles) so that all conditions share a single intensity scale.

\subsection{WMH segmentation methods}
Every condition is segmented by the same four automated WMH segmenters, all recent and widely used: WMH-SynthSeg~\cite{wmhsynthseg}, segcsvd~\cite{segcsvd}, MARS-WMH~\cite{marswmh}, and
TrUE-Net~\cite{truenet}. Each is applied with fixed default settings across all reconstructions and thicknesses, so
that any difference reflects the reconstruction rather than the segmenter. A segmenter can only reveal a reconstruction's effect on lesions it can detect. We therefore benchmark the four
segmenters against the manual reference on the unaltered HR FLAIR, both voxel-wise and per lesion
(Tables~\ref{tab:reader_vs_gt_voxel} and~\ref{tab:reader_vs_gt_detection}), and take as the \emph{primary reader} the
one most sensitive to small lesions on HR (the regime in which erasure and hallucination are decided), so that the
failures we attribute to reconstruction are not the reader's own blindness. By this criterion, we adopt MARS-WMH; the
remaining three are retained to test how far each finding depends on the segmenter.

\subsection{Lesion-level analysis and statistics}
Our analysis relies on two references: the manual mask (true lesions) and the primary reader's segmentation of the unaltered HR FLAIR (its ceiling). This dual-reference approach isolates the marginal effect of the SR reconstruction.
Reference lesions (connected components of the manual mask) were stratified by equivalent diameter (<3, 3--6, 6--10, and >10 mm). A lesion is considered detected if any predicted voxel overlaps it. We report the following metrics:
\begin{enumerate}
\item \textbf{Detection sensitivity:} Stratified by lesion size, using the HR detection rate (Table~\ref{tab:reader_vs_gt_detection}) as the ceiling.

\item \textbf{Voxel-wise accuracy:} Dice, precision, and recall evaluated against the manual reference. Here, 1-recall and 1-precision represent erased and falsely added lesion tissue (the latter capturing the inflation of real lesions). This is summarized as the Dice gain over the un-reconstructed LR baseline (ΔDice; Table~\ref{tab:sr_accuracy_mars}).

\item \textbf{Erasure rate:} The fraction of HR-detected lesions lost post-reconstruction, evaluated by size (Fig.~\ref{fig:err_hal}).

\item \textbf{Hallucination rate:} The number of predicted components per scan that overlap neither the manual mask nor the reader's HR segmentation. This isolates SR-induced false positives from those inherent to the reader (Fig.~\ref{fig:err_hal}).
\end{enumerate}
Uncertainty and paired contrasts were estimated using a subject-clustered bootstrap (5,000 resamples).

\begin{table}[ht]
\centering
\caption{Agreement of four automated WMH segmentation methods with the manual
reference on the 29-subject isotropic cohort,
evaluated on the HR FLAIR.
Values are mean\,$\pm$\,s.d.\ across subjects. Dice, precision and recall are
voxel-wise; HD95 is the 95th-percentile Hausdorff distance;
volume difference is the signed percentage error in total WMH volume relative to
the manual mask.}
\label{tab:reader_vs_gt_voxel}
\begin{tabular}{lccccc}
\toprule
Reader & Dice & Precision & Recall & HD95 (mm) & Volume diff.\ (\%) \\
\midrule
WMH-SynthSeg & 0.35 $\pm$ 0.20 & 0.31 $\pm$ 0.23 & 0.51 $\pm$ 0.17 & 12.3 $\pm$ 5.2 & $+$182 $\pm$ 245 \\
segcsvd      & \textbf{0.63} $\pm$ 0.14 & 0.49 $\pm$ 0.16 & \textbf{0.92} $\pm$ 0.06 & \textbf{5.0} $\pm$ 4.9 & $+$108 $\pm$ 77 \\
MARS-WMH     & \textbf{0.63} $\pm$ 0.21 & 0.67 $\pm$ 0.20 & 0.67 $\pm$ 0.25 & 13.7 $\pm$ 13.9 & \textbf{$+$15} $\pm$ 86 \\
TrUE-Net     & 0.52 $\pm$ 0.17 & \textbf{0.93} $\pm$ 0.07 & 0.37 $\pm$ 0.16 & 16.0 $\pm$ 19.1 & $-$60 $\pm$ 17 \\
\bottomrule
\end{tabular}
\end{table}
\section{Experiments and Results}

\subsection{The segmenters disagree most on the small lesions}
On the unaltered HR FLAIR, the four readers agree moderately with the manual reference and diverge most on the
small lesions that dominate the cohort (Tables~\ref{tab:reader_vs_gt_voxel} and~\ref{tab:reader_vs_gt_detection}).
Sub-3~mm detection sensitivity spans 0.12 to 0.44 across readers, and volumetric bias ranges from severe
over-segmentation to under-segmentation. MARS-WMH is the most sensitive to small lesions, has the highest overall
detection, and is the only reader close to the correct WMH volume, at a Dice matching the best. Because erasure and
hallucination can be measured only through a reader that detects small lesions on HR, we use MARS-WMH as the primary
reader for the reconstruction analysis.

\subsection{Super-resolution recovers lesion signal, with ECLARE leading}
\begin{figure}[htbp]
    \centering
    \includegraphics[width=\linewidth]{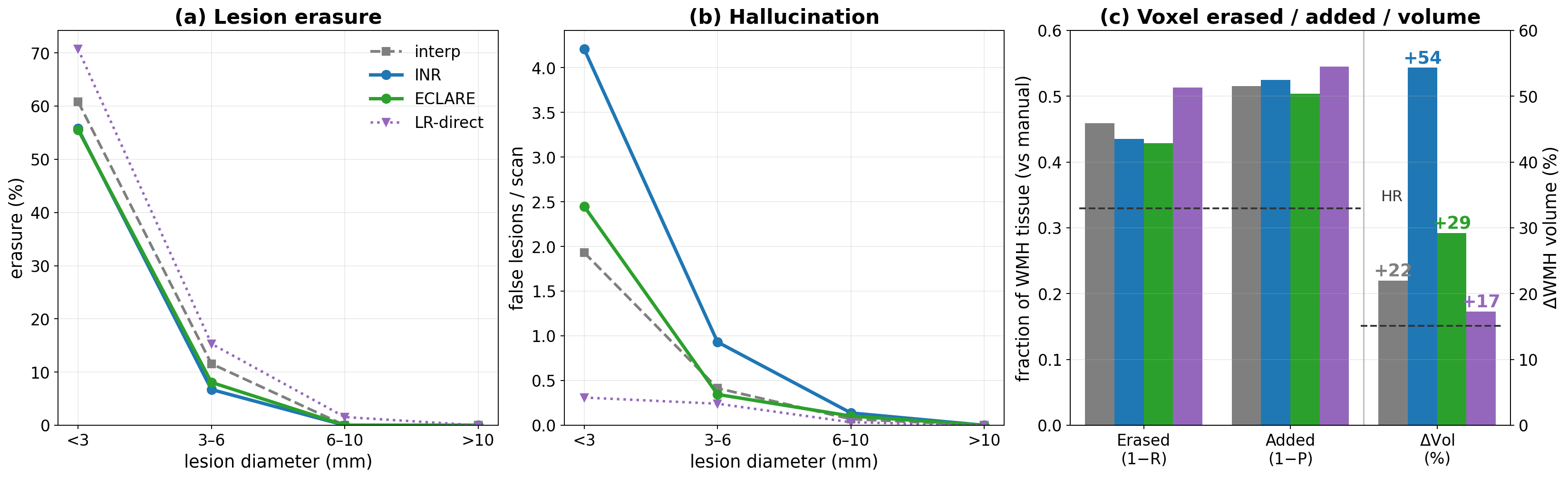}
    \caption{Recon-induced effects at 5 mm (MARS-WMH), the most severe thickness studied; the pattern is milder at 3 mm (Table~\ref{tab:sr_accuracy_mars}). \textbf{(a)} Lesion erasure and \textbf{(b)} hallucination by lesion diameter: (a) percentage of HR-detected manual lesions lost after reconstruction; (b) recon-induced false-positive components per scan. \textbf{(c)} Voxel view vs the manual reference: erased tissue ($1-$recall) and added tissue ($1-$precision) on the left axis, and signed volume difference ($\Delta$Vol) on the right axis; dashed lines mark the HR ceiling. ECLARE adds the least tissue, while INR inflates total WMH volume most ($+$54\%) despite few hallucinated lesions. 29 subjects.}
    \label{fig:err_hal}
\end{figure}

Every reconstruction improved on segmenting the acquired thick slice directly, and the benefit increased with slice
thickness (Table~\ref{fig:err_hal}). ECLARE recovered some lesion signal at both thicknesses (its Dice
gain over the floor was $+$0.064 and $+$0.060 at 3 and 5~mm, the largest of any method). The multi-contrast INR, despite its T1w prior, was no better than cubic interpolation,
so the anatomical prior conferred no measurable benefit for this reader. No reconstruction reached the HR ceiling.

\begin{table}[ht]
\centering
\caption{Lesion-level detection sensitivity of the four segmentation methods
against the manual reference on the HR FLAIR, overall and stratified by lesion
equivalent diameter (29 subjects; 2{,}597 manual lesions). A manual lesion is
counted as detected if any predicted voxel overlaps it.}
\label{tab:reader_vs_gt_detection}
\begin{tabular}{lccccc}
\toprule
Reader & Overall & $<$3\,mm & 3--6\,mm & 6--10\,mm & $>$10\,mm \\
\midrule
WMH-SynthSeg & 0.20 & 0.12 & 0.40 & 0.85 & 1.00 \\
segcsvd      & 0.51 & 0.39 & \textbf{0.87} & \textbf{1.00} & 1.00 \\
MARS-WMH     & \textbf{0.53} & \textbf{0.44} & 0.79 & 0.88 & 1.00 \\
TrUE-Net     & 0.27 & 0.13 & 0.70 & 0.96 & 1.00 \\
\bottomrule
\end{tabular}
\end{table}

\begin{table}[ht]\centering
\caption{SR method comparison for \textbf{MARS-WMH} at 3 and 5\,mm
slice thickness on the 29-subject cohort. D, P
and R are voxel-wise Dice, precision and recall against the manual reference;
$\Delta$D is the Dice gain over the LR-direct at the matching thickness
($\Delta=\text{method}-\text{LR}$; 95\% subject-clustered bootstrap CI). Methods: HR,
high-resolution reference; LR, LR-direct (thick slice, no reconstruction); Interp, cubic interpolation; INR, implicit neural representation; ECLARE, self-supervised single-image SR (slice profile estimated). ECLARE gives the highest Dice and the largest recovery
over LR at both thicknesses. }
\label{tab:sr_accuracy_mars}
\begin{tabular}{l cccc cccc}
\toprule
 & \multicolumn{4}{c}{3\,mm} & \multicolumn{4}{c}{5\,mm} \\
\cmidrule(lr){2-5}\cmidrule(lr){6-9}
Method & D & P & R & $\Delta$D (CI) & D & P & R & $\Delta$D (CI) \\
\midrule
HR & \multicolumn{8}{c}{.631 / .673 / .668\quad (thickness-independent)} \\
\midrule
LR & .522 & .520 & .578 & --- & .454 & .455 & .487 & --- \\
Interp & .565 & .573 & .614 & $+$.043\,(.036,.050) & .491 & .485 & .541 & $+$.038\,(.028,.047) \\
INR & .558 & .540 & \textbf{.640} & $+$.036\,(.023,.047) & .496 & .475 & .565 & $+$.043\,(.025,.058) \\
ECLARE & \textbf{.586} & \textbf{.613} & .626 & \textbf{$+$.064}\,(.052,.073) & \textbf{.514} & \textbf{.497} & \textbf{.571} & \textbf{$+$.060}\,(.048,.071) \\
\bottomrule
\end{tabular}
\end{table}

\subsection{Erasure of small lesions is the dominant failure}
\begin{figure}[htbp]
    \centering
    \includegraphics[width=\linewidth]{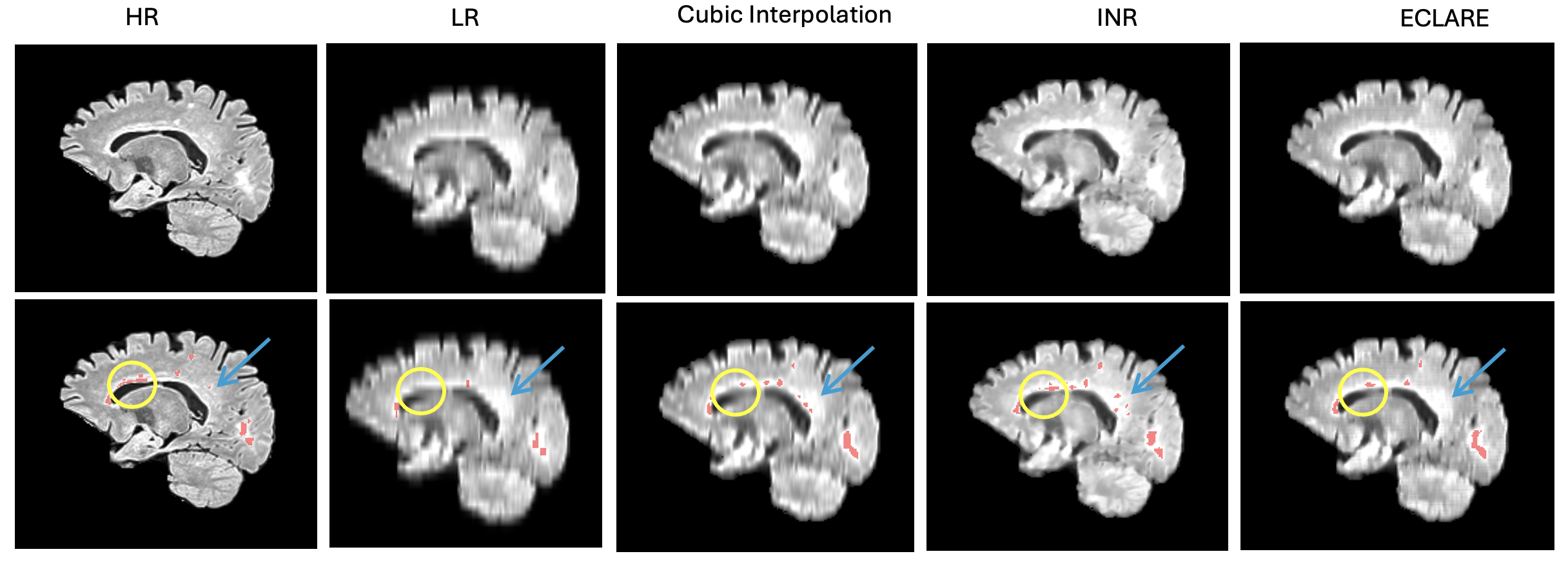}
    \caption{Representative cases (MARS-WMH). Top row: FLAIR under each condition, bottom row: the corresponding MARS-WMH segmentation overlaid. Circles mark erasure; arrows mark hallucination (a component present after reconstruction but absent from both the manual mask and the HR segmentation).}
    \label{fig:cases}
\end{figure}

Decomposing the 5~mm reconstruction error by lesion size (Fig.~\ref{fig:err_hal}) shows the cost is paid overwhelmingly as
erasure of small real lesions rather than as hallucination of false ones. Erasure was concentrated in the smallest
lesions and fell steeply with diameter, whereas reconstruction-induced hallucinations were few at every size. The learned methods erased fewer small
lesions than interpolation or the floor, but INR added more recon-induced false lesions in doing so (Fig.~\ref{fig:err_hal}b); hallucination nonetheless remained the minor failure mode. At the voxel level (Fig.~\ref{fig:err_hal}c) ECLARE adds the least tissue and erases the least, while INR over-segments most by total WMH volume ($+$54\% vs manual), an inflation of real lesions that the lesion count and even the added-tissue fraction understate. At the milder 3~mm thickness, the ordering held with roughly half the erasure (overall 24\% for ECLARE versus 38\% at 5~mm), while INR's volume inflation persisted ($+$51\%), indicating it is a property of the method rather than of slice thickness. Representative cases appear in
Fig.~\ref{fig:cases}, with erased lesions circled and hallucinated components arrowed.

\section{Discussion}

Across 29 isotropic FLAIR scans and a controlled thick-slice simulation, the dominant cost of running SR before WMH segmentation is the erasure of small real lesions, not the hallucination of false ones. Erasure was concentrated in the smallest lesions and increased with slice thickness. No reconstruction restored the high-resolution detection of the small lesions (Fig.~\ref{fig:err_hal}, Table~\ref{tab:sr_accuracy_mars}). Because small, early lesions are often missed, a pipeline that silently removes them reduces the accuracy of tracking WMH over time.

Among the reconstructions, ECLARE achieved the best overall accuracy at both thicknesses, whereas INR, given a T1w as the prior, only matched cubic interpolation under MARS-WMH. The anatomical prior, therefore, did not help recover small WMH because these lesions are often isointense on T1w. A T1-based prior cannot represent the very structure we aim to reconstruct, which explains why the FLAIR-only method performs better in this case. The crucial information came from the learned through-plane recovery based solely on the FLAIR images. INR's parity with interpolation on Dice is moreover misleading: it was the worst reconstruction on the false-positive side, producing the most hallucinations and inflating total WMH volume by over 50\%, an over-segmentation that overlap metrics hide but that directly biases the volume endpoint.

Performance also depends on the downstream segmenter, because only a reader that detects small lesions on HR can reveal whether SR preserves them. The readers we tested bracket a wide range: WMH-SynthSeg, trained by domain randomization, is nearly resolution-invariant, but only because it already misses most small lesions on HR (sub-3 mm sensitivity 0.12), collapsing all conditions toward the same low sensitivity and hiding SR's effect; segcsvd, at the opposite extreme, detects small lesions on HR (0.39) yet erases 70\% of them at 5 mm slice thickness. A resolution-agnostic segmenter only relocates the problem, gaining invariance by missing small lesions. A claim that SR does or does not help WMH segmentation is therefore only interpretable together with the segmenter that produced it.

Because our conclusions rest on a controlled simulation and a single expert reference, they carry several limitations. The low-resolution inputs were simulated from isotropic scans rather than acquired as true thick-slice FLAIR, so reconstruction accuracy is a best-case upper bound that may not transfer to real scanners. Findings also come from a single cohort (ADNI) and one primary segmenter chosen post hoc on the HR scans, so they are conditioned on that segmenter. Finally, because the reference is a single expert's manual mask, some apparent hallucinations may be true lesions it missed; a multi-rater consensus reference would sharpen the false-positive estimates. Future work should validate these findings on real, multi-center anisotropic FLAIR against a multi-rater reference, and on longitudinal WMH change, where erasing small lesions would directly bias progression estimates.

\section{Conclusion}
Applying SR before WMH segmentation does not completely solve the thick-slice problem, its dominant effect is the erasure of small real lesions, not hallucination, and no reconstruction recovers HR sensitivity to the small lesions. This cost is, if anything, understated. Erasure is calculated only for lesions that the primary segmenter detects on HR, and that segmenter identifies fewer than half of sub–3 mm lesions on HR (Table~\ref{tab:reader_vs_gt_detection}). As a result, the ceiling is leaky and the reported erasure rates are conservative. That said, against the realistic baseline of segmenting the thick slice directly, SR is beneficial: the small lesions it recovers outnumber the few thick-slice-visible ones that reconstruction itself removes, so SR reduces, on balance, the false negatives of thick-slice acquisition without reaching HR sensitivity. ECLARE recovers more lesion signal than the T1w-prior INR and interpolation. Whether SR helps at all depends on the WMH segmenter, and SR should be validated for lesion fidelity, not perceptual quality, before it is trusted upstream of WMH quantification.

\bibliographystyle{splncs04}
\bibliography{references}






\end{document}